\documentclass[10pt,twocolumn,letterpaper]{article}

\usepackage{cvpr}              % To produce the CAMERA-READY version
\definecolor{cvprblue}{rgb}{0.21,0.49,0.74}
\usepackage[pagebackref,breaklinks,colorlinks,allcolors=cvprblue]{hyperref}
\usepackage{multirow}
\usepackage{enumitem}
\usepackage{graphicx}
\usepackage{color}
\usepackage{mathtools}
\usepackage{multicol}
\usepackage{xcolor}

\def\paperID{*****} % *** Enter the Paper ID here
\def\confName{CVPR}
\def\confYear{2025}

\title{Tailored-LLaMA: Optimizing Few-Shot Learning in Pruned LLaMA Models with Task-Specific Prompts}

\author{Danyal Aftab\textsuperscript{1} \quad Steven Davy\textsuperscript{2}\\
\textsuperscript{1,2} Technological University Dublin, Ireland\\
{\tt\small D22129961@mytudublin.ie, Steven.Davy@tudublin.ie}
}

\begin{document}
\maketitle
\begin{abstract}
Large language models demonstrate impressive proficiency in language understanding and generation. Nonetheless, training these models from scratch, even the least complex billion-parameter variant demands significant computational resources rendering it economically impractical for many organizations. With large language models functioning as general-purpose task solvers, this paper investigates their task-specific fine-tuning. We employ task-specific datasets and prompts to fine-tune two pruned LLaMA models having 5 billion and 4 billion parameters. This process utilizes the pre-trained weights and focuses on a subset of weights using the LoRA method. One challenge in fine-tuning the LLaMA model is crafting a precise prompt tailored to the specific task. To address this, we propose a novel approach to fine-tune the LLaMA model under two primary constraints: task specificity and prompt effectiveness. Our approach, Tailored LLaMA initially employs structural pruning to reduce the model sizes from 7B to 5B and 4B parameters. Subsequently, it applies a carefully designed prompt specific to the task and utilizes the LoRA method to accelerate the fine-tuning process. Moreover, fine-tuning a model pruned by 50\% for less than one hour restores the mean accuracy of classification tasks to 95.68\% at a 20\% compression ratio and to 86.54\% at a 50\% compression ratio through few-shot learning with 50 shots. Our validation of Tailored LLaMA on these two pruned variants demonstrates that even when compressed to 50\%, the models maintain over 65\% of the baseline model accuracy in few-shot classification and generation tasks. These findings highlight the efficacy of our tailored approach in maintaining high performance with significantly reduced model sizes.
\end{abstract}    
\section{Introduction}
\begin{figure*}[ht]
    \centering
    \includegraphics[width=\linewidth]{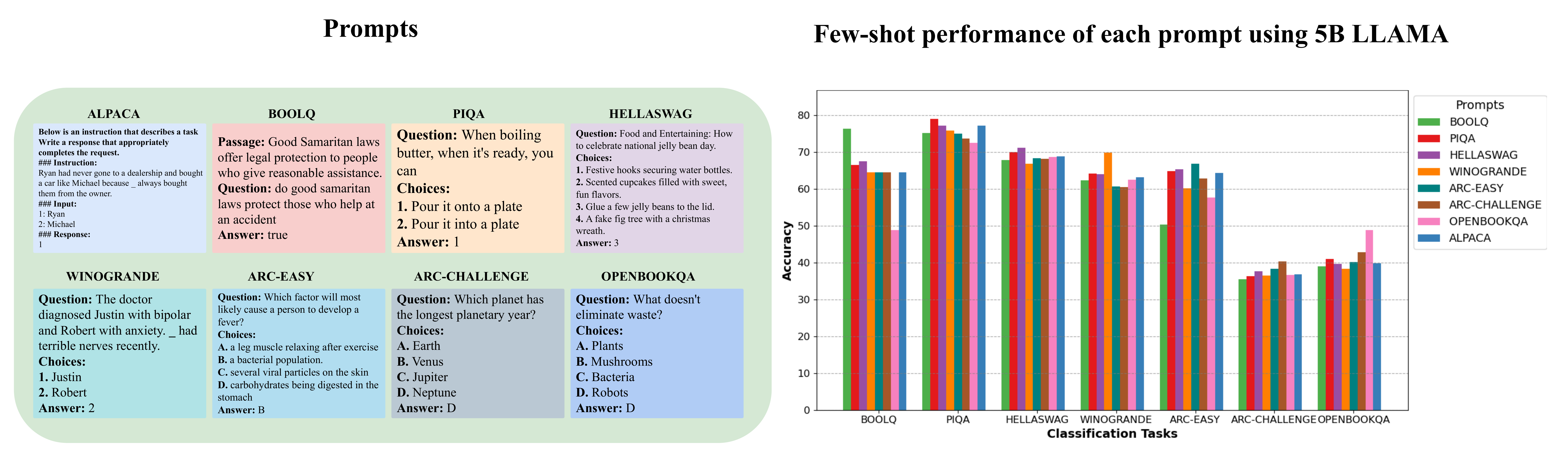}
    \caption{
        A comparative analysis of eight distinct prompts and their respective performance across seven classification tasks. 
    }
    \label{fig:prompts}
\end{figure*}
Large language models (LLMs) \cite{gpt4,bloom,lamda,LLaMA}  trained on massive textual data have demonstrated remarkable proficiency in interpreting complex language-based tasks \cite{gpt-3,palm,chainofthought} and generating text. Consequently, there is a growing interest in developing large-scale language models such as LLaMA \cite{LLaMA}, MPT \cite{mosaicml}, and Falcon \cite{falcon40b} that allow for efficient inference and fine-tuning. These LLMs are available in various sizes each suitable for specific tasks. However, training the LLMs from scratch even for the smallest billion-parameter model requires substantial computational resources which is economically unfeasible for most organizations.

In this paper, we introduce a novel approach to produce a compressed, task-specific, and efficient LLaMA model \cite{LLaMA} by leveraging the pre-trained weights, while having less training cost compared to the one training from scratch. Moreover, we use the structure pruning method to accomplish this objective. Pruning is a widely used method for compressing the task-specific models \cite{deepcompression,hardwarestructuredpruning,blockpruning, pruning,structuredpruning} eliminating redundant parameters to speed up inference while maintaining performance. However, pruning the general purpose LLMs often results in significant performance degradation compared to original models \cite{sparsegpt, llm-pruner,wanda}, especially in scenarios where minimal computational resources are allocated after pruning. In this work, to expedite the fine-tuning process and increase the efficiency of the pruned model under limited data we employ the Low-Rank Adaptation (LoRA) \cite{lora} method.
	
In efficiently fine-tuning the pruned LLaMA model, we identify two primary technical challenges. Firstly, how can we optimize the adaptive weights of a pruned LLaMA model for a specialized task like classification, question-answering, and sentiment analysis? Traditional fine-tuning methods for the sparse LLMs \cite{llm-pruner,structuredpruning} 
depend on datasets designed for multi-tasking approaches. These approaches often result in sub-optimal performance for the specific tasks. Secondly, the selection of appropriate prompts is crucial for attaining optimal performance. Figure \ref{fig:prompts} shows that employing varied prompts across distinct domains results in inconsistent accuracy levels, whereas training with task-specific prompts consistently yields higher accuracy. This demonstrates that even after reducing LLMs with extensive parameters can efficiently adapt to a particular task when fine-tuned with relevant prompts. Our main contributions are:
\begin{itemize}
    \item We propose a novel fine-tuning algorithm for a pruned LLaMA model dubbed targeted task fine-tuning which finetunes a pruned model to a specified target task
    \item We devise a prompt evaluation strategy that selects prompts based on their impact on the task, which enhances the pruned model accuracy and adaptability. This focused approach along with the LoRA method accelerates performance improvement.
    \item We demonstrate the effectiveness of our approach by fine-tuning the LLaMA model across two pruned variants with parameters decreased from 7 billion to 5 billion and 4 billion.
\end{itemize}
Although our experimental focus was on the 7 billion parameter LLaMA model, the Tailored-LLaMA approach exhibits significant potential for generalizability and adaptability to LLMs of varying sizes having fewer parameters than the baseline models.

This paper is organized as follows; Section 2 provides a comprehensive overview of related work in structure pruning and fine-tuning tasks, and Section 3 presents our methodology in detail. Section 4 presents results and experiments, including an extensive ablation study comparing our approach with other fine-tuning methods for pruned LLaMA models.

\section{Related Work}

\paragraph{Network Pruning:} Extensive research has focused on structured pruning as a technique for compressing models in Computer Vision and Natural Language Processing (NLP). This approach is particularly useful for over-parameterized task-specific models such as those used for classification that can sustain significant pruning with minimal loss on performance as evidenced by numerous studies \cite{once,model-compression,deepcompression,dynabert,hardwarestructuredpruning, blockpruning,learning-conv,ai2:winogrande,structured-pruning-llm,learning,structuredpruning}. In contrast, unstructured pruning \cite{model-compression,lotterytickethypothesis,trainbig,movement-pruning} which targets individual neurons rather than entire blocks achieves higher levels of compression but fails to enhance model efficiency making it impractical for accelerating model performance.

In the era of LLMs, the prevailing NLP pipeline has transitioned from specialized models to general-purpose LLMs resulting in limited redundancy. Various approaches such as unstructured pruning, semi-structured pruning \cite{sparsegpt,wanda}, and structured pruning \cite{llm-pruner} have shown a notable performance degradation in LLMs even with moderate sparsity. It is important to note that the aforementioned studies either maintain the original model parameters or tune them minimally. In our work, we view pruning as an initial step and emphasize the need to allocate significant computational resources toward post-structural pruning to regain performance levels.

\paragraph{Transformer language models:} The Transformer model \cite{attention} is a type of architecture that heavily relies on self-attention for sequence-to-sequence tasks. Subsequently, Transformer-based language models have emerged as the leading approach in NLP achieving top performance across various tasks. The introduction of BERT \cite{bert} and GPT-2 \cite{gpt-2} further advanced this field as both are large-scale Transformer language models trained on massive textual data. These new approaches involve fine-tuning the models on specific tasks after pre-training them on general text data leading to significant performance improvements compared to training directly on task-specific data. The ongoing research in this area suggests that training larger Transformer models generally yields better results as evidenced by the continuous development in this direction. GPT-3 \cite{gpt-3} currently holds the record as the largest single Transformer language model having 175 billion parameters.
\begin{figure}[ht]
    % \left
    \includegraphics[width=0.9\linewidth]{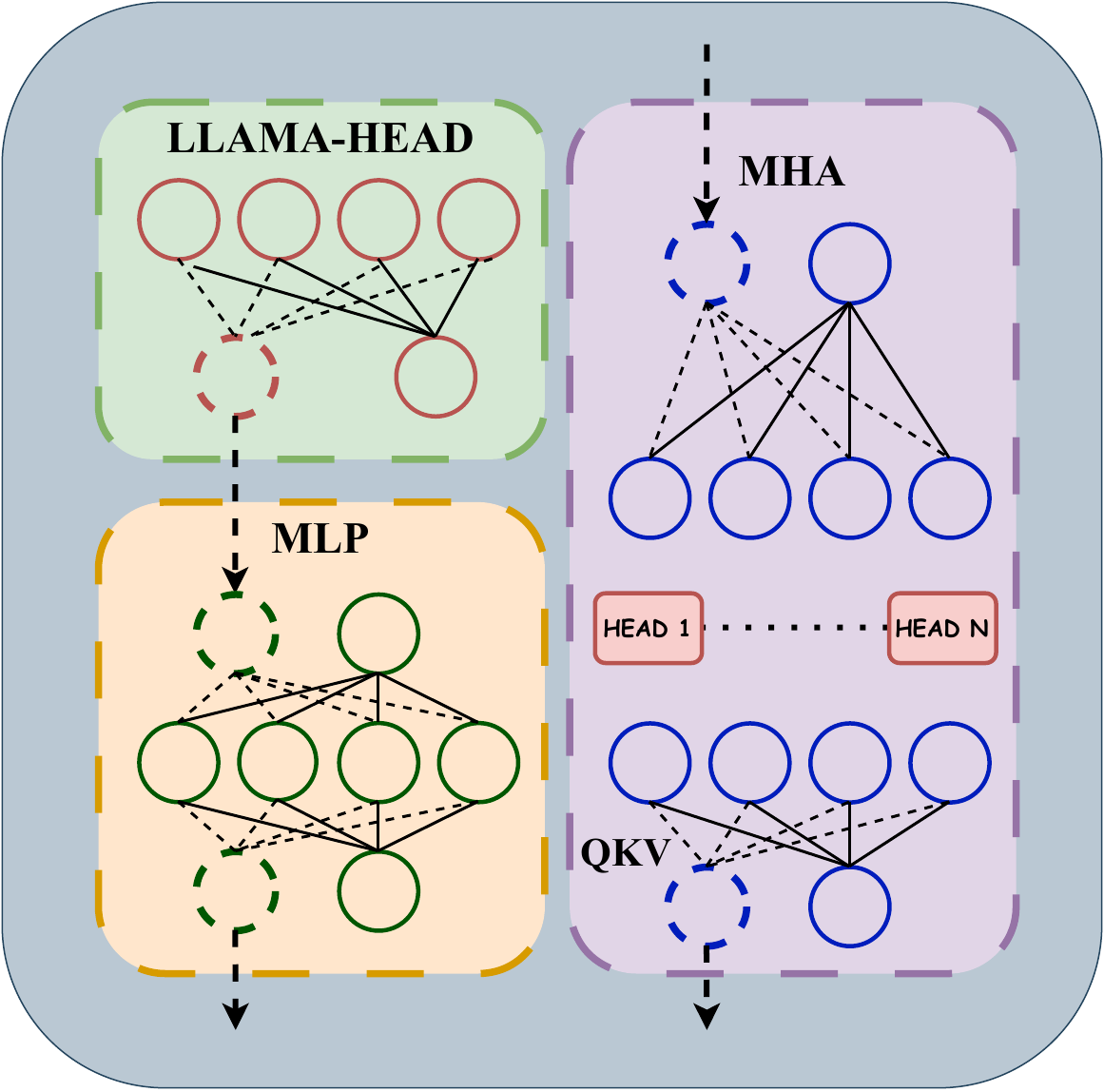}
    \caption{Structural pruning within the LLaMA architecture. The dashed circle in the LLaMA-HEAD represents the trigger neuron initiating the clustering and subsequent pruning of dependent neurons within the MLP and MHA block.
    }
    \label{fig:pruning}
\end{figure}
\paragraph{Prompt Engineering:} While LLaMA 70B \cite{LLaMA} can adjust its behavior with minimal additional training instances, the effectiveness of this adaptation is heavily dependent on the input prompt \cite{gpt-3}. This necessitates the importance of efficiently producing and structuring the prompt to optimize a model performance on a specific task, a practice commonly referred to as prompt engineering. Fine-tuning involves retraining a model initially trained on general datasets for a particular task \cite{bert,radford_improving_nodate}. Various approaches for fine-tuning on a subset of parameters have been proposed \cite{collobert_unified_2008,bert}, however; researchers frequently opt to retrain all parameters to enhance performance for a specific task. Moreover, the vast size of LLaMA 70B poses challenges for traditional fine-tuning methods due to the large amount of checkpoints it generates and the high hardware requirements of having the same memory size compared to the pre-training phase.

%%%%%%%%%%%%%%%%%%%%%%%%%%%%%%%%%%%%%%%%%%%%%%%%%%%%%%%%%%%%%%%%%%%%%%%%

\section{Method}

\begin{table*}[t]
    \centering 
    \resizebox{\linewidth}{!}{
    \begin{tabular}{ll|ccccccc}
        \toprule
        \toprule
         Sparsity & Prompts / Datasets & BoolQ & PIQA & HellaSwag & WinoGrande & ARC-e & ARC-c & OBQA\\
         
        \midrule 
        \multirow{1}{*}{Ratio = 0\%} 
         & - & 76.5 & 79.8 & 76.1 & 70.1 & 72.8 & 47.6 & 57.2  \\
         
        \cmidrule{1-9}
        w/o tune &-&57.06 & 75.68 & 66.80 & 59.83 & 60.94 & 36.52 & 40.0  \\
        \cmidrule{2-9}
        \multirow{8}{*}{\parbox{1.4cm}{Ratio = 20\% \\ w/tune}}

        & Alpaca-Cleaned &64.62 & 77.20 & 68.80 & 63.14 & 64.31 & 36.77 & 39.80 \\
        
        & BoolQ &   \textbf{76.33} & 75.3  & 67.81 & 62.3  & 50.33 & 35.49 & 39 \\

        & PIQA & 66.51 & \textbf{79.00 }& 70    & 64.17 & 64.81 & 36.26 & 41\\
        
        & HellaSwag & 67.52 & 77.26 & \textbf{71.16} & 64.01 & 65.40 & 37.71 & 39.60 \\
        
        & WinoGrande & 64.50 & 75.95 & 66.81 & \textbf{69.96} & 60.19 & 36.43 & 38.40 \\
        
        & ARC-e & 64.49 & 75.02 & 68.40 & 60.70 & \textbf{66.80} & 38.31 & 40.2  \\
        
        & ARC-c & 64.55 & 73.72 & 68.22 & 60.45 & 62.83 & \textbf{43.36} & 42.8  \\
        & OBQA & 48.9  & 72.57 & 68.75 & 62.51 & 57.66 & 36.60 & \textbf{48.8 } \\

        \cmidrule{1-9}
        w/o tune&-& 59.05 & 65.78 & 37.32 & 53.20 & 42.51 & 29.61 & 35.00 \\
        \cmidrule{2-9}
        \multirow{8}{*}{\parbox{1.4cm}{Ratio = 50\% \\ w/tune}} 
        & Alpaca-Cleaned & 59.39 & 71.55 & 55.35 & 57.14 & 51.26 & 30.03 & 37.60 \\
        & BoolQ &\textbf{76.17} & 67.36 & 38.37 & 54.38 & 42.55 & 30.55 & 35.40 \\
        & PIQA &59.88 & \textbf{72.01} & 54.90 & 55.56 & 48.86 & 28.92 & 36.60 \\
        & HellaSwag &59.88 & 71.65 & \textbf{61.70} & 54.85 & 55.43 & 32.08 & 39.80 \\
        
        & WinoGrande & 61.62 & 67.57 & 49.89 & \textbf{61.01} & 44.32 & 31.48 & 36.80 \\

          & ARC-e & 61.63 & 70.89 & 51.72 & 54.22 & \textbf{59.25} & 35.23 & 39 \\
        
        & ARC-c & 62.62 & 70.02 & 51.60& 53.82 & 51.39 & \textbf{36.95} & 39.2  \\
        & OBQA & 61.22 &69.36&50.87&54.93&51.81&34.55&\textbf{42.4} \\
        
        \bottomrule
        \bottomrule
    \end{tabular}
    }
    \caption{Few-shot performance comparison of the pruned LLaMA model post-structural pruning on 8 distinct prompts for 7 classification tasks. The prompt is generated using the identical dataset which is specific to the task. `Bold' indicates the best performance of a prompt within the same task and pruning rate after fine-tuning}
    \label{tab2}
\end{table*}

In this section, we provide a comprehensive description of Tailored-LLaMA. Following the traditional fine-tuning process of pruned LLM models \cite{lora}, Tailored-LLaMA consists of three stages: 
\begin{enumerate}
\item Structure Pruning. This stage focuses on finding the groups of interdependent parameters within LLMs and evaluating their importance to decide which group to prune.
\item Prompt Engineering. This stage involves selecting the most suitable prompt to fine-tune the model.
\item Recovery Phase: Following the selection of the optimal prompt for fine-tuning, this stage proceeds with a fast fine-tuning process employing the LoRA \cite{lora} technique to enhance the model performance efficiently.
\end{enumerate}

\subsection{Structure pruning}
In the context of limited data availability for the post-training process of LLMs, it is imperative to remove the structure inside the model that has minimal impact on model performance when compressing it. This highlights the significance of structure pruning which ensures that interconnected parameters are pruned collectively based on their importance scores. Similar to DepGraph \cite{depgraph}, the dependency graph is built by computing the inter-dependency between layers present in Multi-Head Attention (MHA) and  Feed-Forward Network (FFN) modules. Let $P_i$ and $P_j$ denote two parameters within the model. The terms $\operatorname{In}(P_i)$ and $\operatorname{Out}(P_i)$ refer to all parameters that respectively point toward or point from  $P_i$. The inter-dependency among parameters is defined as shown in Equation (\ref{eq:1}):
\begin{equation}
\resizebox{0.85\linewidth}{!}{%
$P_j \in \operatorname{Out}(P_i)  \wedge \operatorname{Deg}^-(P_j)   = 1 \Rightarrow P_j \text{ is dependent on } P_i$
}
     \label{eq:1}
\end{equation}
Where $\operatorname{Deg}^-(P_j)$ denotes the parameter $P_j$ in-degree, it is important to note that this dependency exhibits direction. Hence, we can correspondingly obtain additional dependency as shown in Equation (\ref{eq:2}):
\begin{equation}
\resizebox{0.85\linewidth}{!}{%
    $ P_i \in \operatorname{In}(P_j) \wedge \operatorname{Deg}^+(P_i)   = 1 \Rightarrow P_i \text{ is dependent on } P_j$
}
\label{eq:2}
\end{equation}
The out-degree of a parameter $P_i$ denoted as $\operatorname{Deg}^+(P_i)$ signifies the number of connections leaving $P_i$. The concept of dependency in this context suggests that if a particular parameter such as Pi relies entirely on another parameter $P_j$ and $P_i$ is pruned then $P_j$ will also need to undergo pruning.

According to the dependency definition, the linked structures in the LLM are evaluated automatically. Any parameter located within the LLM can be regarded as the central initiator possessing the ability to trigger parameters that depend on it. Consequently, these newly activated parameters then act as the subsequent initiators to identify their corresponding parameters that are dependent and activate them. This repetitive process persists until no additional parameters are identified. These identified parameters then form a group for further pruning. Taking LLaMA as an example, this approach analyzes each parameter as the central trigger allowing us to identify all interconnected parameters as illustrated in Figure \ref{fig:pruning}.

To preserve the accuracy of the model, it is important to simultaneously prune the collection of weights in a group. A group denoted by $\mathcal{G} = \{P_i\}_{i=1}^N$ is defined as a set of interconnected parameters, where N is the number of coupled structures in one group and $P_i$ is the weight for each structure.
During the pruning process, the objective is to eliminate the group that has minimal effect on the model predictive performance. This impact can be quantified by analyzing the deviation in the loss function. To assess the specific importance of $P_i$, the change in the loss function can be formulated as Equation (\ref{eq:3}):
\begin{equation}
\resizebox{0.9\linewidth}{!}{%
    $I_{P_i} = | \Delta \mathcal{L}| = |\mathcal{L}_{P_i} - \mathcal{L}_{P_i=0}| =| \underbrace{\frac{\partial \mathcal{L}^{\top}}{\partial P_i} P_i}_{\neq 0}-\frac{1}{2} {P_i}^{\top} H P_i + \mathcal{O}\left(\| P_i \|^3\right) |$
}
\label{eq:3}
\end{equation}
Where $\mathcal{L}$ represents the prediction loss of the next token and $H$ is the hessian matrix. In prior studies \cite{sparsegpt,optimalbraindamage,eigendamage} the initial term denoted as ${\partial \mathcal{L}^{\top}}/{\partial P_i}$ is often disregarded due to the model convergence on training dataset where the gradient of $\mathcal{L}$ concerning $P_i$ is approximately zero. However, as the dataset is not derived from the original training data in this case, the ${\partial \mathcal{L}^{\top}}/{\partial P_i}$ is not close to zero. Since the second term hessian matrix cannot be computed with $\mathcal{O}\left(N^2\right)$ complexity on the LLM, this offers a desired property for determining the significance of $P_i$ by the gradient term under LLMs.
\begin{table*}[!htb]
    \centering 
    \resizebox{\linewidth}{!}{
    \begin{tabular}{ll|cc|ccccccc|cc}
        \toprule
        \toprule
        Sparsity & Method & WikiText2$\color{teal}\downarrow$ & PTB$\color{teal}\downarrow$ & BoolQ & PIQA & HellaSwag & WinoGrande & ARC-e & ARC-c & OBQA & Mean & Recovery Rate \% \\
        \midrule
        
        \multirow{1}{*}{Ratio = 0\%} 
         & LLaMA-7B \cite{LLaMA} & - & - & 76.5 & 79.8 & 76.1 & 70.1 & 72.8 & 47.6 & 57.2 & 68.59 & - \\
         % & LLaMA-7B$^{\star}$ & 12.62 & 22.14 & 73.18 & 78.35 & 72.99 & 67.01 & 67.45 & 41.38 & 42.40 & 63.25 \\
        \cmidrule{1-13}
        \cmidrule{1-13}
        \multirow{6}{*}{\parbox{1.8cm}{Ratio = 20\% \\ Param = 5.4B \\  w/tune }}

        & Wanda \cite{wanda} & 18.43& 33.16& 65.75& 74.70& 64.52& 59.35& 60.65& 36.26& 39.40& 57.23 & 83.43\\
        & FLAP \cite{flap} & \textbf{17.0} & \textbf{30.1} &  69.63 &76.82 &71.20 &68.35 &69.91 &39.25 &39.40 &62.08 & 90.50\\
        & LLM-Pruner \cite{llm-pruner} &  17.58 &30.11 &64.62 &77.20 &68.80 &63.14 &64.31 &36.77 &39.80 &59.23 & 86.35 \\
        & Shortened LLaMA \cite{shortened-LLaMA} &20.2 &32.3 & 75.7& 75.7& \textbf{71.5}& 69.1& 69.9& 41.6& 40.8 & 63.5 & 92.57 \\
        &LoRAPrune \cite{loraprune}&  16.80& 28.75& 65.62& \textbf{79.31}& 70.00& 62.76& 65.87& 37.69& 39.14& 60.05 & 87.55\\
        
        &\textbf{Tailored-LLaMA (Ours)}& 19.09& 34.21& \textbf{76.33}& 79& 71.16& \textbf{69.96} &\textbf{70.80}& \textbf{43.36} &\textbf{48.8}& \textbf{65.63} & \textbf{95.68}\\

        \cmidrule{1-13}
        \multirow{6}{*}{\parbox{1.8cm}{Ratio = 50\% \\ Param = 4.11B \\  w/tune }}

        & Wanda \cite{wanda} &  43.89& 85.87& 50.90& 57.38 &38.12& 55.98& 42.68& 34.20& 38.78& 45.43 & 66.23\\
        & FLAP \cite{flap} & \textbf{29.7} & 53.2  &60.21& 67.52& 52.14& 57.54& 49.66& 29.95& 35.60& 50.37 & 73.44\\
        & LLM-Pruner \cite{llm-pruner} &   38.12& 66.35& 60.28& 69.31 &47.06& 53.43& 45.96& 29.18& 35.60& 48.69 & 70.99 \\
        & Shortened LLaMA \cite{shortened-LLaMA}&  33.2& 58.5&  62.5& 69.2& 60.7& 66.8& 57.4 &34.5& 36.8& 55.4 & 80.83\\
        &LoRAPrune \cite{loraprune}& 30.12& \textbf{50.30}& 61.88& 71.53& 47.86& 55.01& 45.13& 31.62& 34.98& 49.71 & 72.47\\
        
        &\textbf{Tailored-LLaMA (Ours)}& 39.26& 71.96& \textbf{76.17}& \textbf{72.01}& \textbf{61.7}& \textbf{67.01} & \textbf{59.25}& \textbf{36.95} &\textbf{42.4}& \textbf{59.36} & \textbf{86.54} \\
        \bottomrule
        \bottomrule
    \end{tabular}
    }
    \caption{Few-shot performance comparison of the Tailored-LLaMA with other fine-tuning methods post-pruning. The mean and the recovery rate are calculated among seven classification datasets. `Bold' represents the overall best performance within the same compression rate after fine-tuning.}
    \label{tbl:LLaMA_result}
\end{table*}
\begin{table*}[!ht]
    \centering 
    \resizebox{\linewidth}{!}{
    \begin{tabular}{c|cc|ccccccc|c}
        \toprule
        Shots (K) & WikiText2 $\color{teal}\downarrow$ & PTB$\color{teal}\downarrow$ & BoolQ & PIQA & HellaSwag & WinoGrande & ARC-e & ARC-c & OBQA & Average \\
        \midrule

        10 & 19.09 & 34.21 & 67.06 & 75.68 & 66.80 & 68.83 & 60.94 & 38.52 & 44.00 & 60.26 \\
        
        20 & 17.58 & 30.66 & 73.62 & 77.20 & 68.80 & 68.14 & 62.31 & 39.77 & 45.80 & 62.09 \\
        30 & 19.09 & \underline{30.26} & 74.00 & 78.66 & 69.75 & 69.54 & 64.39 & 40.20 & 45.60 & 63.02 \\
        
        40 & 19.39 & 30.57 & 75.24 & \underline{79.00} & 70.52 & 69.85 & 65.48 & 42.01 & 46.00  & 63.73 \\
        
        50 & \underline{17.48} & 70.57 & \underline{76.33} & 78.95 & \underline{71.16} & \underline{69.96} & \underline{66.80} & \underline{43.36} & 47.50 & \underline{64.44} \\
        
        100 & 17.67 & 30.60 & 74.39 & 78.83 & 71.09 & 69.96 & 66.05 & 43.32 & 47.60 & 64.03\\
        
        200 & 17.74 & 30.75 & 75.75 & 78.74 & 70.28 & 69.95 & 66.30 & 43.30 & \underline{48.80} & 64.30 \\      
        \bottomrule
    \end{tabular}
    }
    \caption{The PPL and Accuracy at different shot counts for 20\% compressed LLaMA.}
    \label{overfit_all_datasets}
\end{table*}
The importance of group $\mathcal{G}$ is estimated by aggregating the importance scores of each parameter denoted by $I_{\mathcal{G}} = \sum_{i=1}^{N}I_{P_i}$. After calculating the importance of each group we proceed to assign a rank to each group according to their importance and then prune those with lower importance by a predetermined pruning ratio.

\subsection{Prompt Engineering}
The approach known as reinforcement learning from human feedback (RLHF) \cite{deepreinforcement,RLHF} uses human preferences as a reward signal to fine-tune the LLaMA model and it was used to follow a wide class of textual instructions just like GPT-4. When LLaMA is given a prompt, it initially converts the input text into tokens that the model can understand. These tokens are then processed by transformer layers, which analyze their relationships and context. Within these layers, attention mechanisms assign distinct weights to tokens based on their importance and context. Following the attention process, the model generates its own interpretations of the input data, referred to as intermediate representations. These representations are later transformed back into readable text.

An essential component of this process is the randomness function, which is affected by two key parameters: temperature and top-k sampling. Temperature helps to balance the randomness and predictability of the output. A higher temperature leads to more varied outputs, while a lower temperature results in more predictable outputs. On the other hand, top-k sampling restricts the model choices to the most probable tokens at each stage of output generation. In our approach, we employ the optimal decoding strategy for superior results by using temperature \textbf{1} and top-k sampling \textbf{50}. 

This work shows the effect of prompt on the accuracy of the LLaMA model. Here, we explore a heuristic strategy observed in human reading behavior when they are giving instruction also known as re-reading \cite{re-reading}. When prompted with instructions that lack specificity for the task, the model produces inferior results compared to those generated with task-specific direction as shown in Table \ref{tab2}. Therefore, providing a specific description is crucial for generating precise and relevant outputs.
% Our results indicate that the LLaMA model often generates outputs that are too general when given instructions that are vague or not detailed enough for the specific task as show in Figure 6. 

Effective prompting strategies are crucial for guiding LLMs towards generating desired outputs. This involves formulating clear and specific prompts that minimize ambiguity.  LLM architectures are typically trained on large amounts of textual data encapsulating the combined information from numerous authors. When confronted with a broad or uninformative prompt, the LLM output tends to be generic, applicable in various contexts but potentially sub-optimal for a specific task. Conversely, a detailed and precise prompt reduces the model uncertainty and aligns it towards the appropriate response, enabling the generation of content that aligns more closely with the unique requirements of the given scenario.

\subsection{Recovery Phase}
To recover the accuracy of the model under limited data and expedite the fine-tuning process, it is imperative to select the minimum number of parameters that require updates during the training phase. For this, we utilize the LoRA \cite{lora} method to fine-tune the pruned model. Each parameter denoted as $P$ includes both unpruned and pruned linear projections in the LLM and can be symbolized as $P$. The update value of $\Delta P$ for $P$ can be describe as $\Delta P = RS \in \mathbb{R}^{b^- \times b^+}$, where $R \in \mathbb{R}^{b^- \times b}$ and $S \in \mathbb{R}^{b \times b^+}$. The forward computation can now be represented in Equation (\ref{eq:4}):
\begin{equation}
    f(x) = (P+\Delta P)X + b = (PX + b) + (RS)X
    \label{eq:4}
\end{equation}
Where the bias in the dense layer is represented by $b$, the minus sign $b^-$ and the plus sign $b^+$ distinguish the dimensions of the rows in matrix $R$ from the columns in matrix $S$. By training only the low-rank matrices $R$ and $S$ we achieve a significant reduction in the overall training cost, hence reducing the large amount of data required for training. Furthermore, it is possible to reparameterize the additional parameters $R$ and $S$ into $\Delta P$, thus the final compressed model would not include any extra parameters.

\section{Experiments and results}

\subsection{Dataset and Evaluation}
To demonstrate the effectiveness of Tailored-LLaMA, we test it over two variants of the pruned LLaMA model having 5 billion and 4 billion parameters. We perform few-shot task classification to evaluate the fine-tuned models using lm-evaluation-harness \cite{eval-harness} strategy on common sense reasoning datasets: PIQA~\cite{Bisk2020piqa}, HellaSwag~\cite{zellers2019hellaswag}, BoolQ~\cite{clark-etal-2019-boolq}, WinoGrande~\cite{ai2:winogrande}, ARC-easy~\cite{allenai:arc}, ARC-challenge~\cite{allenai:arc} and OpenbookQA~\cite{OpenBookQA2018}. Furthermore, we supplement our evaluation with a few-shot perplexity (PPL) analysis on two language modeling datasets WikiText2~\cite{merity2016pointer} and PTB~\cite{marcus-etal-1993-building}.

\subsection{Implementation Details}
During the model pruning process 20 samples were arbitrarily chosen from Bookcorpus \cite{Zhu_2015_ICCV} and reduced to a sequence length of 128 to compute the gradient. For the recovery stage, we employ few-shot learning using task-specific datasets. For instance, the Hellaswag dataset was used for the Hellaswag task, the PIQA dataset was employed for the PIQA task, and so on. Remarkably, tuning these models on average requires less than 1 hour on a single GPU with only 3 epochs. We follow the same strategy as LoRA \cite{lora} for fine-tuning. We set the rank $d$ to 8 and a learning rate to 1e-4 with 100 warming steps. The training batch size is 64 and the AdamW optimizer is used for our experiment. We found 3 epochs best for training the model among 1 to 6 as increasing the number of epochs had a negative impact on the model performance. We conduct our experiment on A100 single GPU having 80GB of memory for approximately 0.8 hours.

\subsection{Few-shot performance}
We evaluate the few-shot performance of a fine-tuned LLaMA model on two pruned variants: 6 billion and 5 billion parameters as shown in Table \ref{tab2}. Our analysis demonstrates that fine-tuning the pruned LLaMA model on a task-specific dataset consistently yields better performance compared to training on a composite dataset. For instance, the LLaMA model pruned to a sparsity ratio of 20\% achieved an accuracy of 76.33 when fine-tuned on the BoolQ dataset. This performance surpassed its accuracy on other fine-tuning datasets, including PIQA, HellaSwag, WinoGrande, ARC-e, ARC-c, and OBQA. Similarly, fine-tuning the LLaMA model pruned by 50\% yields an accuracy of 72.01 on the PIQA task which surpasses its performance on all other datasets used in the fine-tuning process. These patterns are consistently observed across 7 tasks, as indicated by the bold values, which suggest that employing a dataset aligned with the specific task remarkably benefits the pruned model performance. Additionally, in the case of the 50\% pruned LLaMA model, the BoolQ prompt achieved an accuracy of 76.17, which is 99.57\% of the baseline accuracy of 76.5, thereby surpassing other prompts in restoring performance. Furthermore, the  PIQA, HellaSwag, WinoGrande, ARC-e, ARC-c, and OBQA prompts preserved $90.24\%$, $81.08\%$, $95.59\%$, $81.38\%$, $77.62\%$ and $74.12\%$ of their original performance. Despite fine-tuning the pruned LLaMA model for less than \textbf{1 hour}, our Tailored-LLaMA outperforms other prominent fine-tuning methods by achieving a mean recovery rate of $\textbf{95.68\%}$ for compression ratio $\textbf{20\%}$ and $\textbf{86.54\%}$ for $\textbf{50\%}$ post-structural pruning of comparable scales as shown in Table \ref{tbl:LLaMA_result}. This signifies the feasibility of using the Tailored-LLaMA to effectively fine-tune the LLaMA model within a short period.
\subsection{Ablation Study}
We conduct tests on all proposed prompts mentioned in Figure \ref{fig:prompts}. The results can be found in Table \ref{tab2}. To learn the specific representation of each task we conduct an ablation study for various $K$ shots as shown in Table \ref{overfit_all_datasets}.
Our findings from Table \ref{overfit_all_datasets} suggest an upward trend in performance with an increase in sample size indicating that the larger datasets generally enhance model capabilities. However, this improvement is not strictly linear and shows dataset-specific variances. For instance, while the Accuracy for BoolQ and PPL for PTB remains relatively stable across sample sizes, the accuracy for HellaSwag and WinoGrande improves more noticeably. Surprisingly, the accuracy for ARC-e increases at 200 shots after a decrease at 100 shots, suggesting a non-linear relationship between sample size and model performance. The average performance across all datasets trends upward, although with slight fluctuations, underscoring the fact that while additional data can be beneficial it does not guarantee proportional enhancements in model accuracy, and each dataset interacts differently with the sample size. This nuanced behavior suggests that the model learning and generalization capacity is significantly influenced by the nature and diversity of the dataset, a point of consideration for the adaptability and scalability of the compressed LLaMA model in various contexts.  As a result, the capacity of the model to generalize to novel data could potentially be impaired. The empirical evidence from our experiments indicates that employing a set of 50 shots is optimal for enhancing the training process and achieving maximal accuracy.

\section{Conclusion}
This paper proposes a novel method for constructing a compressed, task-specific, and efficient LLaMA model by leveraging domain-specific prompts. This approach offers a more cost-effective solution compared to training a LLaMA on a composite dataset. Firstly, we accomplish structure pruning by iteratively analyzing each parameter within the model as a central trigger to construct dependency groups, thereby constructing the LLaMA dependency graph. Subsequently, we evaluate the significance of these groups using parameter-wise estimation. Secondly, we fine-tune the LLaMA model using task-specific datasets and prompts. Lastly, to reduce the recovery time of LLaMA we use the LoRA method. 
We evaluate the efficacy of Tailored-LLaMA on two pruned LLaMA models with capacities of 5 billion and 4 billion parameters, using multiple few-shot datasets. Our experimental results indicate that Tailored-LLaMA outperforms other prominent fine-tuning methods.

\section{Acknowledgements}
This publication has emanated from research conducted with the financial support of Science Foundation Ireland under Grant number 21/FFP-A/9174
{
    \small
    \bibliographystyle{ieeenat_fullname}
    \bibliography{main}

\begin{thebibliography}{51}
\providecommand{\natexlab}[1]{#1}
\providecommand{\url}[1]{\texttt{#1}}
\expandafter\ifx\csname urlstyle\endcsname\relax
  \providecommand{\doi}[1]{doi: #1}\else
  \providecommand{\doi}{doi: \begingroup \urlstyle{rm}\Url}\fi

\bibitem[Almazrouei et~al.(2023)Almazrouei, Alobeidli, Alshamsi, Cappelli, Cojocaru, Debbah, Goffinet, Hesslow, Launay, Malartic, et~al.]{falcon40b}
Ebtesam Almazrouei, Hamza Alobeidli, Abdulaziz Alshamsi, Alessandro Cappelli, Ruxandra Cojocaru, Merouane Debbah, Etienne Goffinet, Daniel Hesslow, Julien Launay, Quentin Malartic, et~al.
\newblock The falcon series of open language models.
\newblock \emph{arXiv preprint arXiv:2311.16867}, 2023.

\bibitem[An et~al.(2024)An, Zhao, Yu, Tang, and Wang]{flap}
Yongqi An, Xu Zhao, Tao Yu, Ming Tang, and Jinqiao Wang.
\newblock Fluctuation-based adaptive structured pruning for large language models.
\newblock In \emph{Proceedings of the AAAI Conference on Artificial Intelligence}, pages 10865--10873, 2024.

\bibitem[Bisk et~al.(2020)Bisk, Zellers, Bras, Gao, and Choi]{Bisk2020piqa}
Yonatan Bisk, Rowan Zellers, Ronan~Le Bras, Jianfeng Gao, and Yejin Choi.
\newblock Piqa: Reasoning about physical commonsense in natural language.
\newblock In \emph{Thirty-Fourth AAAI Conference on Artificial Intelligence}, 2020.

\bibitem[Brown et~al.(2020)Brown, Mann, Ryder, Subbiah, Kaplan, Dhariwal, and Neelakantan]{gpt-3}
Tom~B. Brown, Benjamin Mann, Nick Ryder, Melanie Subbiah, Jared Kaplan, Prafulla Dhariwal, and Neelakantan.
\newblock Language models are few-shot learners.
\newblock \emph{arXiv:2005.14165 [cs]}, 2020.

\bibitem[Cai et~al.(2019)Cai, Gan, Wang, Zhang, and Han]{once}
Han Cai, Chuang Gan, Tianzhe Wang, Zhekai Zhang, and Song Han.
\newblock Once-for-all: Train one network and specialize it for efficient deployment.
\newblock In \emph{International Conference on Learning Representations}, 2019.

\bibitem[Chowdhery et~al.(2023)Chowdhery, Narang, Devlin, Bosma, Mishra, Roberts, Barham, Chung, Sutton, Gehrmann, et~al.]{palm}
Aakanksha Chowdhery, Sharan Narang, Jacob Devlin, Maarten Bosma, Gaurav Mishra, Adam Roberts, Paul Barham, Hyung~Won Chung, Charles Sutton, Sebastian Gehrmann, et~al.
\newblock Palm: Scaling language modeling with pathways.
\newblock \emph{Journal of Machine Learning Research}, 24\penalty0 (240):\penalty0 1--113, 2023.

\bibitem[Christiano et~al.(2017)Christiano, Leike, Brown, Martic, Legg, and Amodei]{deepreinforcement}
Paul~F Christiano, Jan Leike, Tom Brown, Miljan Martic, Shane Legg, and Dario Amodei.
\newblock Deep reinforcement learning from human preferences.
\newblock \emph{Advances in neural information processing systems}, 30, 2017.

\bibitem[Clark et~al.(2019)Clark, Lee, Chang, Kwiatkowski, Collins, and Toutanova]{clark-etal-2019-boolq}
Christopher Clark, Kenton Lee, Ming-Wei Chang, Tom Kwiatkowski, Michael Collins, and Kristina Toutanova.
\newblock {B}ool{Q}: Exploring the surprising difficulty of natural yes/no questions.
\newblock In \emph{Proceedings of the 2019 Conference of the North {A}merican Chapter of the Association for Computational Linguistics: Human Language Technologies, Volume 1 (Long and Short Papers)}, pages 2924--2936, Minneapolis, Minnesota, 2019. Association for Computational Linguistics.

\bibitem[Clark et~al.(2018)Clark, Cowhey, Etzioni, Khot, Sabharwal, Schoenick, and Tafjord]{allenai:arc}
Peter Clark, Isaac Cowhey, Oren Etzioni, Tushar Khot, Ashish Sabharwal, Carissa Schoenick, and Oyvind Tafjord.
\newblock Think you have solved question answering? try arc, the ai2 reasoning challenge.
\newblock \emph{arXiv:1803.05457v1}, 2018.

\bibitem[Collobert and Weston(2008)]{collobert_unified_2008}
Ronan Collobert and Jason Weston.
\newblock A unified architecture for natural language processing: deep neural networks with multitask learning.
\newblock In \emph{Proceedings of the 25th international conference on {Machine} learning}, pages 160--167, New York, NY, USA, 2008. Association for Computing Machinery.

\bibitem[Deng et~al.(2020)Deng, Li, Han, Shi, and Xie]{model-compression}
Lei Deng, Guoqi Li, Song Han, Luping Shi, and Yuan Xie.
\newblock Model compression and hardware acceleration for neural networks: A comprehensive survey.
\newblock \emph{Proceedings of the IEEE}, 108\penalty0 (4):\penalty0 485--532, 2020.

\bibitem[Devlin et~al.(2019)Devlin, Chang, Lee, and Toutanova]{bert}
Jacob Devlin, Ming-Wei Chang, Kenton Lee, and Kristina Toutanova.
\newblock Bert: Pre-training of deep bidirectional transformers for language understanding.
\newblock In \emph{North American Chapter of the Association for Computational Linguistics}, 2019.

\bibitem[Fang et~al.(2023)Fang, Ma, Song, Mi, and Wang]{depgraph}
Gongfan Fang, Xinyin Ma, Mingli Song, Michael~Bi Mi, and Xinchao Wang.
\newblock Depgraph: Towards any structural pruning.
\newblock In \emph{Proceedings of the IEEE/CVF Conference on Computer Vision and Pattern Recognition}, pages 16091--16101, 2023.

\bibitem[Frankle and Carbin(2018)]{lotterytickethypothesis}
Jonathan Frankle and Michael Carbin.
\newblock The lottery ticket hypothesis: Finding sparse, trainable neural networks.
\newblock In \emph{International Conference on Learning Representations}, 2018.

\bibitem[Frantar and Alistarh(2023)]{sparsegpt}
Elias Frantar and Dan Alistarh.
\newblock Sparsegpt: Massive language models can be accurately pruned in one-shot.
\newblock In \emph{International Conference on Machine Learning}, pages 10323--10337. PMLR, 2023.

\bibitem[Gao et~al.(2023)Gao, Tow, Abbasi, Biderman, Black, DiPofi, Foster, Golding, Hsu, and Le~Noac'h]{eval-harness}
Leo Gao, Jonathan Tow, Baber Abbasi, Stella Biderman, Sid Black, Anthony DiPofi, Charles Foster, Laurence Golding, Jeffrey Hsu, and Alain Le~Noac'h.
\newblock A framework for few-shot language model evaluation, 2023.

\bibitem[Han et~al.(2016)Han, Mao, Dally, and Dally]{deepcompression}
Song Han, Huizi Mao, Dally, and William Dally.
\newblock Deep compression: Compressing deep neural networks with pruning, trained quantization and huffman coding.
\newblock In \emph{International Conference on Learning Representations}, 2016.

\bibitem[Hou et~al.(2020)Hou, Huang, Shang, Jiang, Chen, and Liu]{dynabert}
Lu Hou, Zhiqi Huang, Lifeng Shang, Xin Jiang, Xiao Chen, and Qun Liu.
\newblock Dynabert: Dynamic bert with adaptive width and depth.
\newblock \emph{Advances in Neural Information Processing Systems}, 33:\penalty0 9782--9793, 2020.

\bibitem[Hu et~al.(2022)Hu, Shen, Wallis, Allen-Zhu, Li, Wang, Wang, and Chen]{lora}
Edward~J Hu, Yelong Shen, Phillip Wallis, Zeyuan Allen-Zhu, Yuanzhi Li, Shean Wang, Lu Wang, and Weizhu Chen.
\newblock Lo{RA}: Low-rank adaptation of large language models.
\newblock In \emph{International Conference on Learning Representations}, 2022.

\bibitem[Kim et~al.(2024)Kim, Kim, Kim, Castells, Choi, Shin, and Song]{shortened-LLaMA}
Bo-Kyeong Kim, Geonmin Kim, Tae-Ho Kim, Thibault Castells, Shinkook Choi, Junho Shin, and Hyoung-Kyu Song.
\newblock Shortened llama: A simple depth pruning for large language models.
\newblock \emph{ICLR Workshop on Mathematical and Empirical Understanding of Foundation Models (ME-FoMo)}, 2024.

\bibitem[Kurtic et~al.(2023)Kurtic, Frantar, and Alistarh]{hardwarestructuredpruning}
Eldar Kurtic, Elias Frantar, and Dan Alistarh.
\newblock Ziplm: Hardware-aware structured pruning of language models.
\newblock \emph{arXiv preprint arXiv:2302.04089}, 2023.

\bibitem[Lagunas et~al.(2021)Lagunas, Charlaix, Sanh, and Rush]{blockpruning}
Fran{\c{c}}ois Lagunas, Ella Charlaix, Victor Sanh, and Alexander~M Rush.
\newblock Block pruning for faster transformers.
\newblock \emph{arXiv preprint arXiv:2109.04838}, 2021.

\bibitem[LeCun et~al.(1989)LeCun, Denker, and Solla]{optimalbraindamage}
Yann LeCun, John Denker, and Sara Solla.
\newblock Optimal brain damage.
\newblock In \emph{Advances in Neural Information Processing Systems}. Morgan-Kaufmann, 1989.

\bibitem[Li et~al.(2016)Li, Kadav, Durdanovic, Samet, and Graf]{pruning}
Hao Li, Asim Kadav, Igor Durdanovic, Hanan Samet, and Hans~Peter Graf.
\newblock Pruning filters for efficient convnets.
\newblock In \emph{International Conference on Learning Representations}, 2016.

\bibitem[Li et~al.(2020)Li, Wallace, Shen, Lin, Keutzer, Klein, and Gonzalez]{trainbig}
Zhuohan Li, Eric Wallace, Sheng Shen, Kevin Lin, Kurt Keutzer, Dan Klein, and Joey Gonzalez.
\newblock Train big, then compress: Rethinking model size for efficient training and inference of transformers.
\newblock In \emph{International Conference on machine learning}, pages 5958--5968. PMLR, 2020.

\bibitem[Liu et~al.(2017)Liu, Li, Shen, Huang, Yan, and Zhang]{learning-conv}
Zhuang Liu, Jianguo Li, Zhiqiang Shen, Gao Huang, Shoumeng Yan, and Changshui Zhang.
\newblock Learning efficient convolutional networks through network slimming.
\newblock In \emph{Proceedings of the IEEE international conference on computer vision}, pages 2736--2744, 2017.

\bibitem[Ma et~al.(2023)Ma, Fang, and Wang]{llm-pruner}
Xinyin Ma, Gongfan Fang, and Xinchao Wang.
\newblock Llm-pruner: On the structural pruning of large language models.
\newblock \emph{Advances in neural information processing systems}, 36:\penalty0 21702--21720, 2023.

\bibitem[Marcus et~al.(1993)Marcus, Santorini, and Marcinkiewicz]{marcus-etal-1993-building}
Mitchell~P. Marcus, Beatrice Santorini, and Mary~Ann Marcinkiewicz.
\newblock Building a large annotated corpus of {E}nglish: The {P}enn {T}reebank.
\newblock \emph{Computational Linguistics}, 19\penalty0 (2):\penalty0 313--330, 1993.

\bibitem[Merity et~al.(2017)Merity, Xiong, Bradbury, and Socher]{merity2016pointer}
Stephen Merity, Caiming Xiong, James Bradbury, and Richard Socher.
\newblock Pointer sentinel mixture models.
\newblock In \emph{International Conference on Learning Representations}, 2017.

\bibitem[Mihaylov et~al.(2018)Mihaylov, Clark, Khot, and Sabharwal]{OpenBookQA2018}
Todor Mihaylov, Peter Clark, Tushar Khot, and Ashish Sabharwal.
\newblock Can a suit of armor conduct electricity? a new dataset for open book question answering.
\newblock In \emph{EMNLP}, 2018.

\bibitem[OpenAI(2023)]{gpt4}
OpenAI.
\newblock Gpt-4 technical report.
\newblock \emph{ArXiv}, abs/2303.08774, 2023.

\bibitem[Radford et~al.(2018)Radford, Narasimhan, Salimans, Sutskever, et~al.]{radford_improving_nodate}
Alec Radford, Karthik Narasimhan, Tim Salimans, Ilya Sutskever, et~al.
\newblock Improving language understanding by generative pre-training.
\newblock \emph{ArXiv}, 2018.

\bibitem[Radford et~al.(2019)Radford, Wu, Child, Luan, Amodei, Sutskever, et~al.]{gpt-2}
Alec Radford, Jeffrey Wu, Rewon Child, David Luan, Dario Amodei, Ilya Sutskever, et~al.
\newblock Language models are unsupervised multitask learners.
\newblock \emph{OpenAI blog}, 1\penalty0 (8):\penalty0 9, 2019.

\bibitem[Sakaguchi et~al.(2021)Sakaguchi, Bras, Bhagavatula, and Choi]{ai2:winogrande}
Keisuke Sakaguchi, Ronan~Le Bras, Chandra Bhagavatula, and Yejin Choi.
\newblock Winogrande: An adversarial winograd schema challenge at scale.
\newblock \emph{Communications of the ACM}, 64\penalty0 (9):\penalty0 99--106, 2021.

\bibitem[Sanh et~al.(2020)Sanh, Wolf, and Rush]{movement-pruning}
Victor Sanh, Thomas Wolf, and Alexander Rush.
\newblock Movement pruning: Adaptive sparsity by fine-tuning.
\newblock \emph{Advances in Neural Information Processing Systems}, 33:\penalty0 20378--20389, 2020.

\bibitem[Scao et~al.(2022)Scao, Fan, Akiki, Pavlick, Ili{\'c}, Hesslow, Castagn{\'e}, Luccioni, Yvon, Gall{\'e}, et~al.]{bloom}
Teven~Le Scao, Angela Fan, Christopher Akiki, Ellie Pavlick, Suzana Ili{\'c}, Daniel Hesslow, Roman Castagn{\'e}, Alexandra~Sasha Luccioni, Fran{\c{c}}ois Yvon, Matthias Gall{\'e}, et~al.
\newblock Bloom: A 176b-parameter open-access multilingual language model.
\newblock \emph{arXiv preprint arXiv:2211.05100}, 2022.

\bibitem[Stiennon et~al.(2020)Stiennon, Ouyang, Wu, Ziegler, Lowe, Voss, Radford, Amodei, and Christiano]{RLHF}
Nisan Stiennon, Long Ouyang, Jeffrey Wu, Daniel Ziegler, Ryan Lowe, Chelsea Voss, Alec Radford, Dario Amodei, and Paul~F Christiano.
\newblock Learning to summarize with human feedback.
\newblock \emph{Advances in Neural Information Processing Systems}, 33:\penalty0 3008--3021, 2020.

\bibitem[Sun et~al.(2024)Sun, Liu, Bair, and Kolter]{wanda}
Mingjie Sun, Zhuang Liu, Anna Bair, and J~Zico Kolter.
\newblock A simple and effective pruning approach for large language models.
\newblock In \emph{The Twelfth International Conference on Learning Representations}, 2024.

\bibitem[Team et~al.(2023)]{mosaicml}
MN Team et~al.
\newblock Introducing mpt-7b: A new standard for open-source, commercially usable llms, 2023.
\newblock \emph{URL www. mosaicml. com/blog/mpt-7b. Accessed}, pages 05--05, 2023.

\bibitem[Thoppilan et~al.(2022)Thoppilan, De~Freitas, Hall, Shazeer, Kulshreshtha, Cheng, Jin, Bos, Baker, Du, et~al.]{lamda}
Romal Thoppilan, Daniel De~Freitas, Jamie Hall, Noam Shazeer, Apoorv Kulshreshtha, Heng-Tze Cheng, Alicia Jin, Taylor Bos, Leslie Baker, Yu Du, et~al.
\newblock Lamda: Language models for dialog applications.
\newblock \emph{arXiv preprint arXiv:2201.08239}, 2022.

\bibitem[Touvron et~al.(2023)Touvron, Lavril, Izacard, Martinet, Lachaux, Lacroix, Rozi{\`e}re, Goyal, Hambro, Azhar, et~al.]{LLaMA}
Hugo Touvron, Thibaut Lavril, Gautier Izacard, Xavier Martinet, Marie-Anne Lachaux, Timoth{\'e}e Lacroix, Baptiste Rozi{\`e}re, Naman Goyal, Eric Hambro, Faisal Azhar, et~al.
\newblock Llama: Open and efficient foundation language models.
\newblock \emph{arXiv preprint arXiv:2302.13971}, 2023.

\bibitem[Vaswani et~al.(2017)Vaswani, Shazeer, Parmar, Uszkoreit, Jones, Gomez, Kaiser, and Polosukhin]{attention}
Ashish Vaswani, Noam Shazeer, Niki Parmar, Jakob Uszkoreit, Llion Jones, Aidan~N Gomez, {\L}ukasz Kaiser, and Illia Polosukhin.
\newblock Attention is all you need.
\newblock In \emph{Proceedings of the 31st International Conference on Neural Information Processing Systems}, pages 6000--6010, 2017.

\bibitem[Wang et~al.(2019)Wang, Grosse, Fidler, and Zhang]{eigendamage}
Chaoqi Wang, Roger Grosse, Sanja Fidler, and Guodong Zhang.
\newblock Eigendamage: Structured pruning in the kronecker-factored eigenbasis.
\newblock In \emph{International conference on machine learning}, pages 6566--6575, 2019.

\bibitem[Wang et~al.(2020)Wang, Wohlwend, and Lei]{structured-pruning-llm}
Ziheng Wang, Jeremy Wohlwend, and Tao Lei.
\newblock Structured pruning of large language models.
\newblock In \emph{Proceedings of the 2020 Conference on Empirical Methods in Natural Language Processing (EMNLP)}, pages 6151--6162, 2020.

\bibitem[Wei et~al.(2022)Wei, Wang, Schuurmans, Bosma, Xia, Chi, Le, Zhou, et~al.]{chainofthought}
Jason Wei, Xuezhi Wang, Dale Schuurmans, Maarten Bosma, Fei Xia, Ed Chi, Quoc~V Le, Denny Zhou, et~al.
\newblock Chain-of-thought prompting elicits reasoning in large language models.
\newblock \emph{Advances in neural information processing systems}, 35:\penalty0 24824--24837, 2022.

\bibitem[Wen et~al.(2016)Wen, Wu, Wang, Chen, and Li]{learning}
Wei Wen, Chunpeng Wu, Yandan Wang, Yiran Chen, and Hai Li.
\newblock Learning structured sparsity in deep neural networks.
\newblock \emph{Advances in neural information processing systems}, 29, 2016.

\bibitem[Xia et~al.(2022)Xia, Zhong, and Chen]{structuredpruning}
Mengzhou Xia, Zexuan Zhong, and Danqi Chen.
\newblock Structured pruning learns compact and accurate models.
\newblock In \emph{Proceedings of the 60th Annual Meeting of the Association for Computational Linguistics (Volume 1: Long Papers)}, pages 1513--1528, Dublin, Ireland, 2022. Association for Computational Linguistics.

\bibitem[Xu et~al.(2023)Xu, Tao, Shen, Xu, Xu, Long, and Lou]{re-reading}
Xiaohan Xu, Chongyang Tao, Tao Shen, Can Xu, Hongbo Xu, Guodong Long, and Jian-guang Lou.
\newblock Re-reading improves reasoning in language models.
\newblock arXiv:2309.06275, 2023.

\bibitem[Zellers et~al.(2019)Zellers, Holtzman, Bisk, Farhadi, and Choi]{zellers2019hellaswag}
Rowan Zellers, Ari Holtzman, Yonatan Bisk, Ali Farhadi, and Yejin Choi.
\newblock Hellaswag: Can a machine really finish your sentence?
\newblock In \emph{Proceedings of the 57th Annual Meeting of the Association for Computational Linguistics}, 2019.

\bibitem[Zhang et~al.(2023)Zhang, Chen, Shen, Yang, Ou, Yu, and Zhuang]{loraprune}
Mingyang Zhang, Hao Chen, Chunhua Shen, Zhen Yang, Linlin Ou, Xinyi Yu, and Bohan Zhuang.
\newblock Loraprune: Pruning meets low-rank parameter-efficient fine-tuning.
\newblock \emph{arXiv preprint arXiv:2305.18403}, 2023.

\bibitem[Zhu et~al.(2015)Zhu, Kiros, Zemel, Salakhutdinov, Urtasun, Torralba, and Fidler]{Zhu_2015_ICCV}
Yukun Zhu, Ryan Kiros, Rich Zemel, Ruslan Salakhutdinov, Raquel Urtasun, Antonio Torralba, and Sanja Fidler.
\newblock Aligning books and movies: Towards story-like visual explanations by watching movies and reading books.
\newblock In \emph{The IEEE International Conference on Computer Vision (ICCV)}, 2015.

\end{thebibliography}
}

% WARNING: do not forget to delete the supplementary pages from your submission 
% \input{sec/X_suppl}

\end{document}